# Intelligent Car System

**Qasim Siddique**

**Department of Computer Science, Foundation University
Islamabad, Pakistan**
qasim_1987@hotmail.com

**ABSTRACT.** In modern life the road safety has becomes the core issue. One single move of a driver can cause horrifying accident. The main goal of intelligent car system is to make communication with other cars on the road. The system is able to control to speed, direction and the distance between the cars the intelligent car system is able to recognize traffic light and is able to take decision according to it. This paper presents a framework of the intelligent car system. I validate several aspect of our system using simulation.

## 1 Introduction

The percentage of the road accident is increasing day by day. Most of the innocent people lost their life in the road accident so there have always been great needs of intelligent systems. Recent advance in the artificial intelligent suggest autonomous vehicle navigation and communication will be possible in the near future. Individual car can be equipped with feature of autonomy such as cruise control.

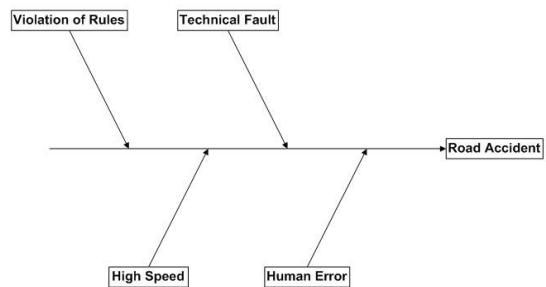

**Figure 1 Causes of Road Accident**

There are many causes of the road accident but most of them are due to the human error. The human error can be avoided by using intelligent system in the vehicle making communication between the vehicles can make the journey safer. Intelligent systems in the vehicles will help to secure the lives of people on the roads, prevent traffic jams and to decrease the car accidents.

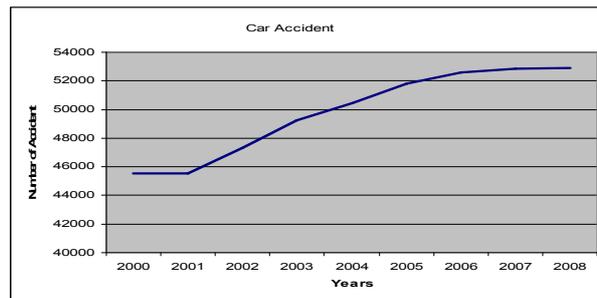

**Figure 2 Number of Car Accident in New South Wales**

The intelligent car system will increased the people life security on the road. It will keep track of safety related information about the vehicle, its passengers and its cargo. Intelligent car system will recognize the traffic signals. Intelligent car system the traffic signal and take the decision accordion to it. Intelligent car system will deliver fast accurate and complete travel information to help traveler to decide whether to make trip.

This paper is organized as fellows: Section 2 Present the Architecture of intelligent car system Section 3 Present the Recognize of traffic light by the intelligent car system Section 4 Present How the Car will make communication between the other cars in Section 5 Result are presented and finally paper summery are given in last section.

## 2. Car Architecture

This section will explain the architecture of the intelligent car system.

Figure 3 explains the architecture of the intelligent car system. The intelligent car system takes the input from sensor, Camera and Satellite. The intelligent car system gets the input from these devices. Then the decision is taken based on these inputs.

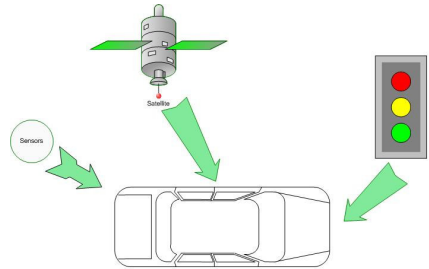

**Figure 3 Architecture of intelligent car System**

Sensors help the intelligent car system to know the distance of the other object from the car. If the object is near the car will indicate the user to stop else it will let the car to move. The satellite helps the intelligent car system to decide the route which is better, safer and faster. The information retrieve from the satellite contain the route map and the current position of the cars. The traffic signal information is given to the intelligent car system by the help of the camera. These images are then passed to the intelligent vision system the intelligent vision system decides that the car should move forward or it should stay in the current position. The intelligent vision system contains a vision algorithm.

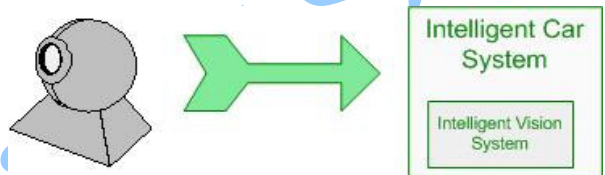

**Figure 4 Architecture of Intelligent Vision System**

The figure 4 explains the architecture of the intelligent vision system. In the intelligent car system there is a vision system which perform the analysis on the image taken by the camera. In perform the decision in the real time environment.

## Basic Hardware Architecture of Car

I required the following hardware for testing of the intelligent car

- Toy car

Containing battery and two motors (clock and anti clock wise direction).motor will help the car to move forward and backwards and left and right.

- IR Sensor

The four sensors (IR Sensor) will be required in the car. Which will be placed in the front, back, left and right direction in the car (the main purpose of the sensor is to detect the object in it surrounding)

- Microcontroller (8051)

All the components of the car will be connected to the microcontroller. All four sensor of the car will be connected to the Port 1 of the microcontroller (10-17pins). It is used to active the sensor when the car wants to change it direction. The motor will be connected in to the port 2 of the microcontroller.

- Front DC motor

The front dc motor will be connected to the front wheels of the car to change its direction left or right

- Back DC motor

The back Dc motor will be connected to the back wheel of the car.

- Sound Speaker

This is used to produce sound when a specific interrupt occur.

- Oscillator Module

This is used to give clock pulse to the microcontroller

- Antenna

This will make the communication with other cars

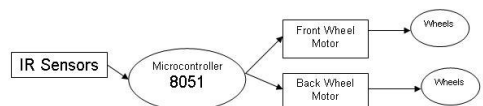

**Figure 5 Basic working structure of car**

## 3. Recognize of Traffic Light

The section explains how the traffic light is recognizing by the intelligent car system.

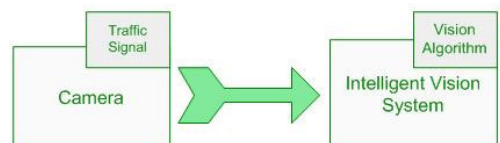

**Figure 6 Data Flow from Camera to Intelligent Vision control System**

There are two basic components of the intelligent vision system one the vision algorithm and the other is camera. The camera gives the image of the real world traffic signal which is on the road side. The camera then passes this image to the intelligent vision system. Then the vision algorithm takes this image from the memory of intelligent vision system. Then the vision algorithm performs the analysis on the images. The main goal of the vision algorithm is to detect the traffic signal then the pass this information to the intelligent system so that the intelligent car system can take the decision according to the requirements.

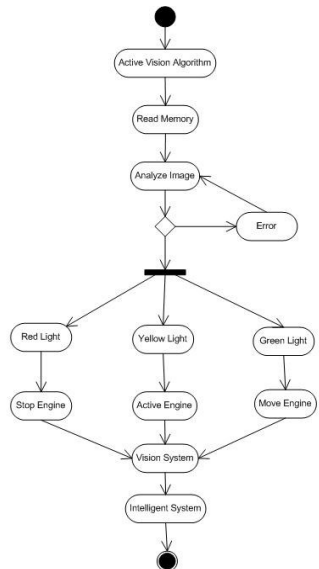

**Figure 7 Activity of Vision Algorithm**

When the vision algorithm is active it read the memory of the intelligent vision system. Then it performs the analysis on the images and detects the color of the signal which is glowing at that time after it has detected the color of the traffic signal. Then it selects the decision and then passes the decision according to that signal. This signal is then passed to the vision system and then the intelligent vision system passes the information to the intelligent car system so that the information could to transfer to the engine.

## A.Signal Selection

- RED light

Turn the Engine in the stop mode.

- YELLOW light

Turn the Engine in the Active mode but the engine should not be moving.
- GREEN light

Turn the Engine in the Moving mode.

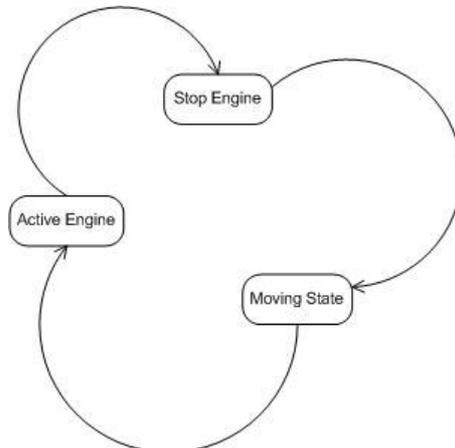

**Figure 8 Engine States**

# B. Engine State

- Stop Engine

The engine is not moving and is turned off
- Moving State

The Engine is ON and is moving
- Active Engine

The Engine is ON but it is not moving

## 4. Car to Car Communication

This section explains the car to car communication. The car to car communication helps the driver to understand the movement of the other driver. The intelligent car system contains a communication model which helps the driver of the car to make the communication with the other car.

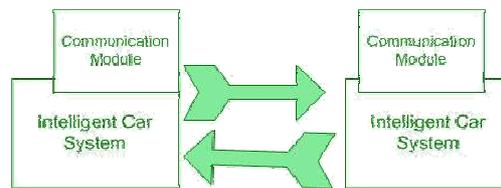

**Figure 9 Communication Module of Intelligent Car System**

Communication module is also an other important part of the intelligent car system is keep on searching the other car system and connect with them when they found each other. The communication module have a specific radius as it only connect with the car which are within is radius.

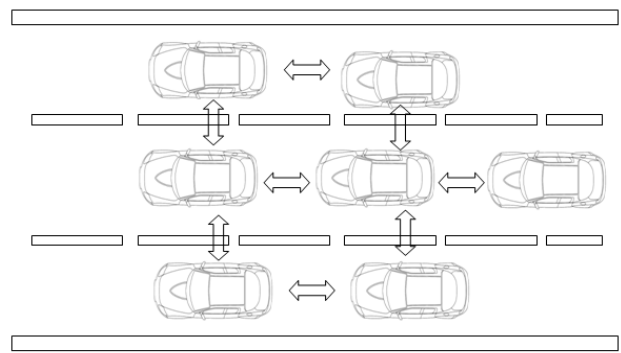

**Figure 10 How Car Make Communication with Each other**

The above figure 10 show that how the car will make the communication with each car in the above fig is sending the message to the other car to find out the next move of each other.

## A. Scenarios

- Other vehicle suddenly came in front of intelligent car system

As it happened in our everyday life that during driving a vehicle suddenly came in front and we have to apply the brake.

But in case of intelligent car the car will have to perform some calculation. Will calculate distance from that object. Will calculate the distance of that object which is in the back direction of the intelligent car if the distance of both the object from the intelligent car is enough then the car will apply brake.

- Distance not enough

In this case the distance at both the object are not enough from the intelligent car then the car have to calculate the distance from it left side and right side and where ever it finds the required space the car will move to that side.

- How it will manage speed and brake

Whenever there is more distance from the object in front of the car the car will accelerate more. Whenever there is less distance from the object in front of the car the car will have to decrease speed.

## B. Table Management in Intelligent car System

Ever intelligent car system will have three main tables in there memory on which they will change make or change the decision table:

- Network Table
- Present state and direction table
- Decision state and direction table

**1) Network Table**

The network table will contain the information of the entire cars which is nearest to the intelligent car

**2) Present state and direction table**

This table will contain the present states of the car and the present direction of the car.

**3) Decision state and direction table**

This table will contain the next state decision of the car and the next direction of the car. All the entries in the network table will be updated from this table. This table entry will be build by the help of destination point and network table.

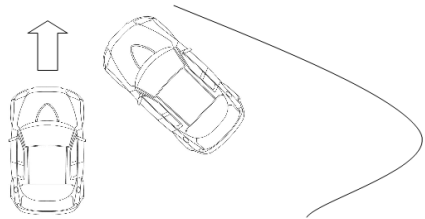

**Figure 11 Avoiding Accident**

As there will be a proper communication system between every car on the road so this will reduce the chance of the road accident.

## 5. Results

I evaluate various aspect of the intelligent car system using simulation. In the simulation various components of the intelligent car system was tested in different scenario with the simulation testing I obtained the following results. The main goal of designing the simulation is to calculate the accuracy of the main components of the intelligent car system with is as followed:

- Decision Support System
- Communication module
- Intelligent Vision System
- I/O Units of Intelligent Car System

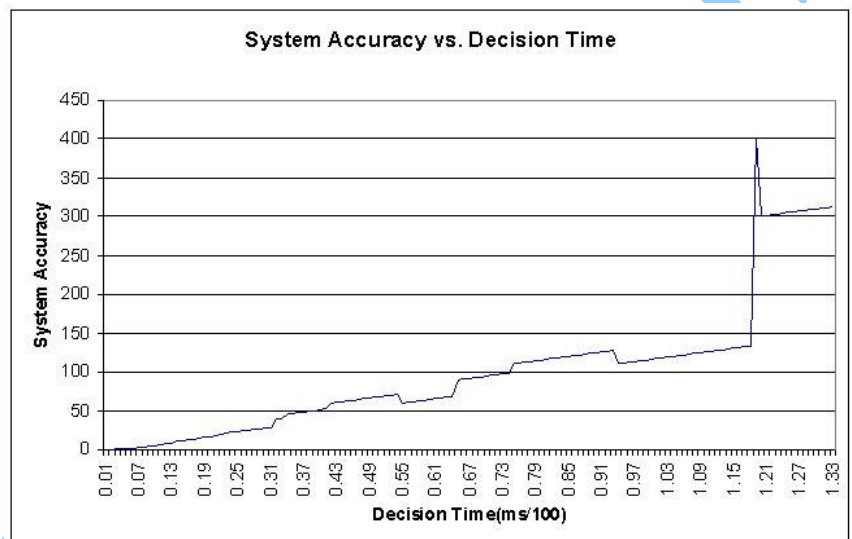

**Figure 12 System Accuracy vs. Decision Time**

The above graph explains the relationship between the System accuracy and decision time of the intelligent car system. The X-axis of the graph displays the decision time and the y-Axis of the graph display the System accuracy of the intelligent car system. It can be observed from the graph as the time increased the accuracy of the system also increased.

The Figure 13 graph explains the relationship between the system accuracy and the number of cars. The System accuracy is on the y-axis of the graph and the numbers of cars are on the x-axis of the graph. It can be observed from the graph as the number of the car increased the system accuracy also increased which help in avoiding the traffics accidents.

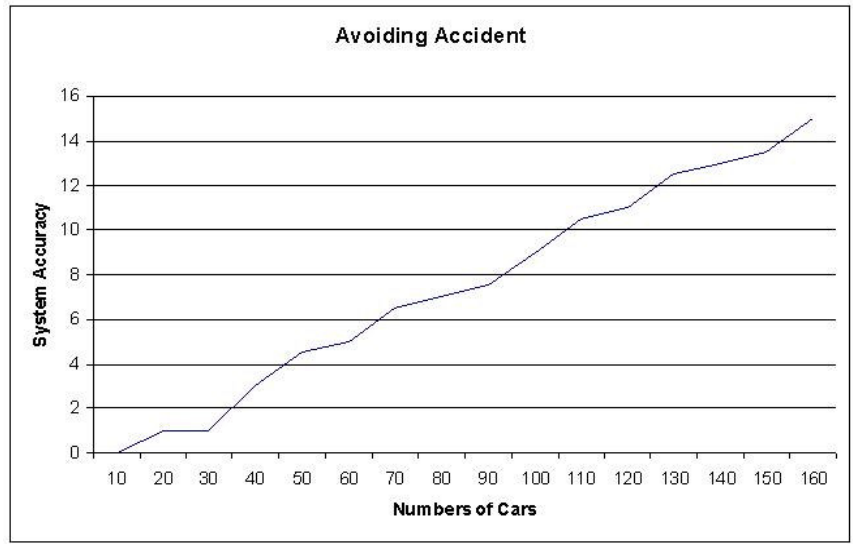

**Figure 13 Avoiding Accident**

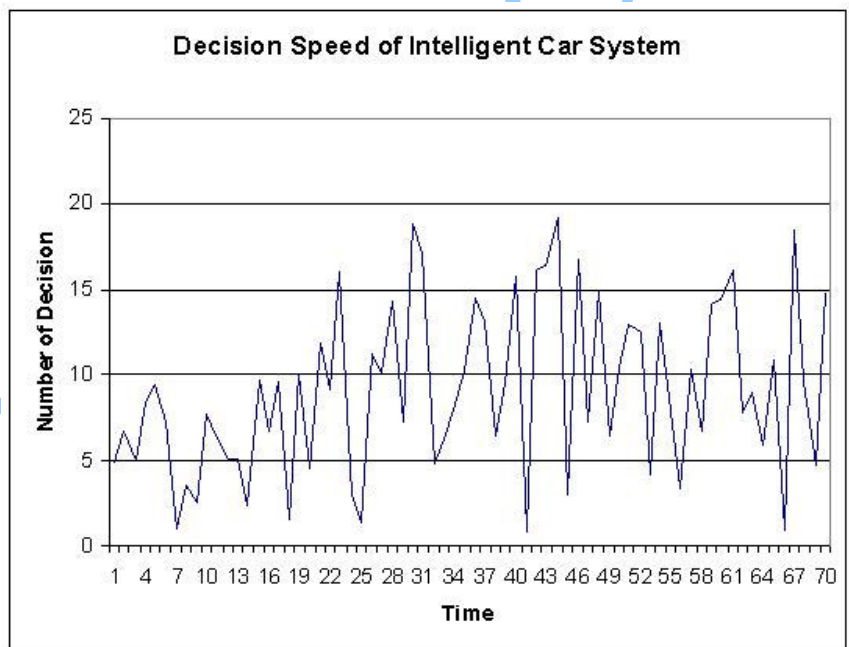

**Figure 14 Decision Speed of intelligent Car System**

The above graph explains the relationship between the Number of decision and Time of the intelligent car system. The X-axis of the graph display the Time and the y-Axis of the graph display the number of Decision of the intelligent car system. At different interval of time the intelligent car system takes different decisions.

## Summary

This paper presents the architecture of the intelligent car system. The intelligent car system consist different components which will help the drives to overcome the road accident and make the life of the paper safer on the road. The deployment of the intelligent car system will help to avoid the car accident and it will also help in avoiding traffic jams. Intelligent vision system will help to recognize the traffic signal.

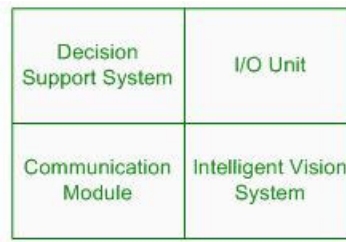

**Figure 15 Main Components of Intelligent Car System**

## Acknowledgments

I am deeply indebted to my teacher Muaz A. Niazi, for his continuous support, encouragement and guidance through my research. Muaz A. Niazi provided valuable insights and a lot of help during my graduate career.

I would like to thank my family and friends specially Saqlian Altaf Nasib for their love, care and support. I would not have accomplished anything without them.